\pdfoutput=1

\documentclass[11pt]{article}

\usepackage{emnlp2021}

\usepackage{times}
\usepackage{latexsym}
\usepackage{makecell}
\usepackage[T1]{fontenc}
\usepackage{booktabs}
\usepackage{graphicx}

\usepackage[utf8]{inputenc}

\usepackage{microtype}

\title{Contrastive Language–Image Pre-training for the Italian Language}

\author{Federico Bianchi \\  Bocconi University \\  Milan, Italy  \\ \textit{f.bianchi@unibocconi.it}
        \And  Giuseppe Attanasio \\ Politecnico di Torino \\  Turin, Italy \\ \textit{giuseppe.attanasio@polito.it}
        \And
        Raphael Pisoni \\ Independent Researcher \\ Vienna, Austria \\ 
        \textit{raphael.pisoni@gmail.com}
        \AND 
        Silvia Terragni \\ University of Milano-Bicocca \\ Milan, Italy \\ \textit{s.terragni4@campus.unimib.it}\And  
        Gabriele Sarti \\ University of Groningen \\Groningen, The Netherlands \\ \textit{g.sarti@rug.nl} \And
        Sri Lakshmi \\ Independent Researcher \\ Chennai, India \\ \textit{srilakshmiimhskalirs@gmail.com}  \\}

\begin{document}

\maketitle
\begin{abstract}
CLIP (Contrastive Language–Image Pre-training) is a very recent multi-modal model that jointly learns representations of images and texts. The model is trained on a massive amount of English data and shows impressive performance on zero-shot classification tasks. Training the same model on a different language is not trivial, since data in other languages might be not enough and the model needs high-quality translations of the texts to guarantee a good performance. 
In this paper, we present the first CLIP model for the Italian Language (CLIP-Italian), trained on more than 1.4 million image-text pairs. Results show that CLIP-Italian outperforms the multilingual CLIP model on the tasks of image retrieval and zero-shot classification.
\end{abstract}

\section{Introduction}

The recent interest in combining different kinds of source domains to incorporate broader context in the training process has led to a surge in multi-modal models spanning modalities like text and vision~\cite{CLIP21} or text and speech~\cite{Schneider2019wav2vecUP}. A multi-modal architecture learns by jointly optimizing its state and parameters on two or more input domains (e.g., images, texts, tabular data, audio signals), with a cost function that may vary depending on the task.

Contrastive Language–Image Pre-training (CLIP)~\cite{CLIP21} is among the most recent multi-modal models for joint learning representations of image and text. The neural network proposed by OpenAI learns to pair visual concepts to their description in natural language by leveraging a contrastive loss that pushes images and their respective captions closer in the embedding space. CLIP is thus trained on a large-scale dataset composed of images and their corresponding captions. The dataset used in this context contains 400 million images collected on the web.

Although the model shows impressive zero-shot performance at many supervised tasks, its capabilities are bounded to the language on which the model is trained, i.e. the English language. Although there have been efforts to introduce multilingual CLIP models,\footnote{\url{https://huggingface.co/sentence-transformers/clip-ViT-B-32-multilingual-v1}} it is well-known that multilingual models tend not to perform as well as language-specific ones~\cite{nozza2020mask}.

We thus propose to fine-tune a specialized version of CLIP on a language different than English, i.e. Italian. Indeed, in this paper, we describe CLIP-Italian\footnote{The choice of Italian comes from the nationality of some of the authors. However, the approach presented in this paper can be generalized to other language and domain text-image pairs without loss of generality.}, a model we developed during the Hugging Face Community Week.\footnote{\url{https://discuss.huggingface.co/t/open-to-the-community-community-week-using-jax-flax-for-nlp-cv/7104}} In the spirit of transparency, we release our best performing model\footnote{\url{https://huggingface.co/clip-italian/clip-italian}} along with
all the produced material. Appendix~\ref{sec:appendix} provides additional details on such releases.

\paragraph{Contributions.} We combine different data sources to generate the largest publicly available multi-modal dataset for the Italian language, with 1.4M image-caption pairs. We use this dataset to train and release the first CLIP image-text model for the Italian language. Moreover, we provide an online demo that can be used to test the model's capabilities. Finally, we show that this model performs better than its multi-lingual counterpart in two well-established multi-modal tasks, namely image retrieval and zero-shot image classification.

\section{Contrastive Language–Image Pre-training}
The task on which CLIP is trained is as follows: given an image in input, the model has to predict which of the different texts is the correct element to associate with the input image. Although it is not mandatory, the correct element typically contains a textual description of the content of the image. In this way, the model learns to associate visual concepts and their respective natural language description.

CLIP's architecture consists of two distinct encoders, one for images and one for texts. The two encoders share the same embedding space. At training time, all the images and texts of a given a mini-batch are projected to a 512 dimensional space. Next, vector similarities are computed for each pair of image and text embeddings and cross entropy loss is applied. The average loss along the image and text dimensions is the final contrastive loss.

\begin{figure*}
    \centering
    \includegraphics[width=0.65\textwidth]{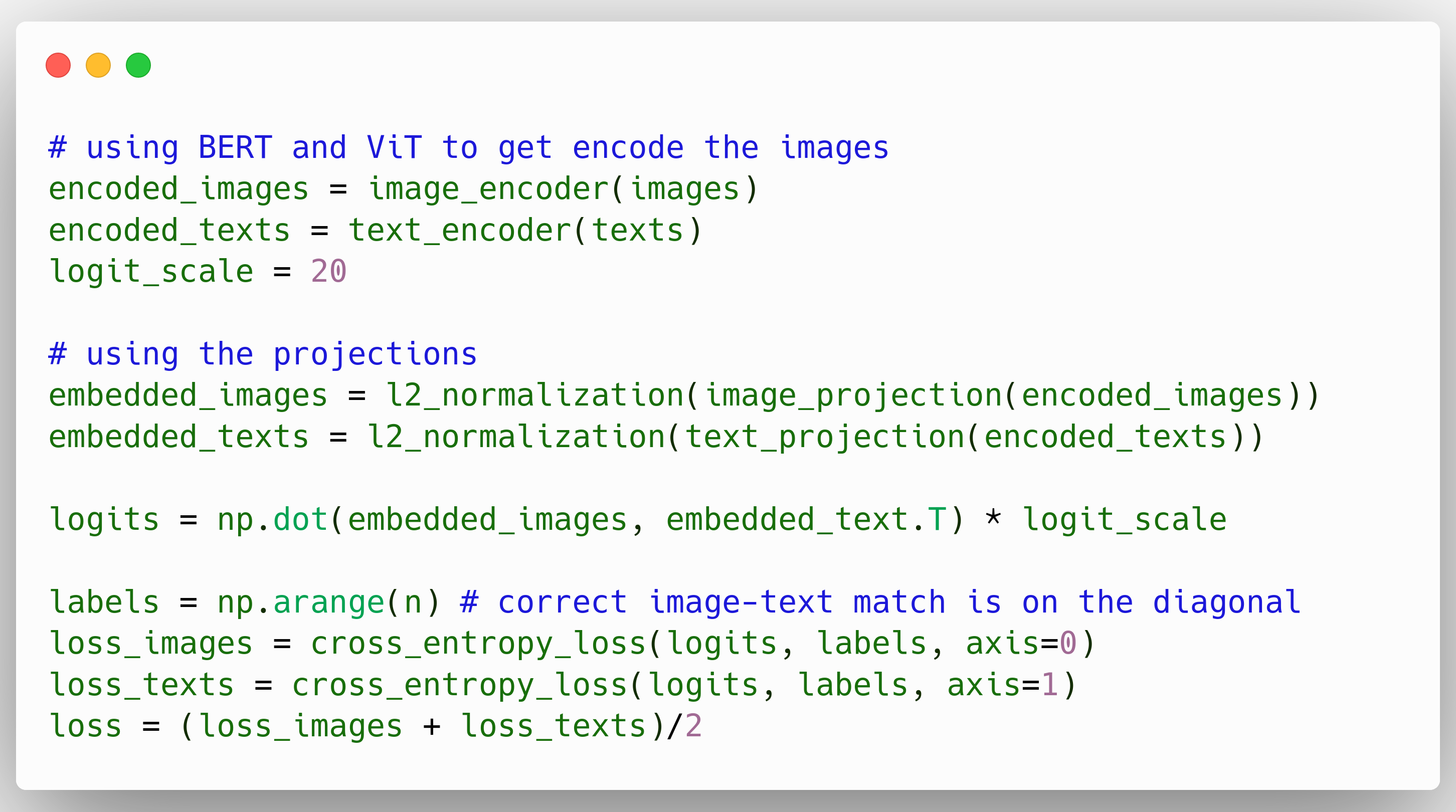}
    \caption{Numpy-like pseudo code that describes the CLIP loss, image is inspired to the one provided by~\cite{CLIP21}.}
    \label{fig:base:configuration}
\end{figure*}

Figure~\ref{fig:base:configuration} briefly summarizes the training process and the loss employed by CLIP to learn the representations.

After training, CLIP can be used for a variety of different tasks: since it embeds image and text in the same space it is possible to tackle image retrieval and zero-shot classification of images by looking at the similarities between texts and images.

Our version of CLIP differs from the original. Mainly, we do not pre-train the encoders from scratch but we start from previous checkpoints for both the vision and the text components. The advantage of this approach is that we use models that are already tuned and we just need to apply a language-specific fine-tuning of the components to create CLIP-Italian. We extensively cover training details in Section~\ref{sec:training}.

\section{Datasets}

We hereby describe the four different sources of data we have used to train our CLIP-Italian model.

\begin{itemize}
    \item  WIT~\cite{srinivasan2021wit} is an image-caption dataset collected from Wikipedia. It is a multilingual dataset, but we pre-processed and extracted the Italian subset. For each image, there are different possible captions. In this work we focused on the Reference Description captions, as they were described in the paper as the most topical and highest quality captions.

    \item MSCOCO-IT~\cite{mscocoit2019}.\footnote{\url{https://github.com/crux82/mscoco-it}} The captions of this image-caption dataset come from the original MSCOCO dataset~\cite{lin2014mscoco} and translated with Microsoft Translator. The 2017 version of the MSCOCO training set contains more than 100K images. More than one caption is available for each image.

    \item Conceptual Captions (CC)~\cite{sharma2018conceptual}.\footnote{\url{https://github.com/google-research-datasets/conceptual-captions}} In this dataset, there are more than 3 millions image-caption pairs, collected from the web. We downloaded the images with the URLs provided by the dataset, but some of them were not available. Eventually, we translated approximately 710K captions to Italian using DeepL.\footnote{\url{https://www.deepl.com/}} 
    
    \item La Foto del Giorno (ILPOST).\footnote{\url{https://www.ilpost.it/foto-del-giorno/}} This image-caption dataset is collected from \textit{Il Post}, a prominent Italian online newspaper. Starting from early 2011, every day, the editors at Il Post have been selecting several images picturing the most salient events in the world. Each photo comes along with an Italian caption. The resulting collection contains almost 30K pairs of images-captions. 

\end{itemize}

\begin{table}[!t]
\centering
\begin{tabular}{@{}lr@{}}
\toprule
Dataset & Captions \\ \midrule
WIT & 525,950 \\
MSCOCO-IT & 116,195 \\
CC & 712,890 \\
ILPOST & 29,055 \\ \bottomrule
\end{tabular}
\caption{The number of captions collected per dataset.}
\label{tab:datasets}
\end{table}

\subsection{Translations}

Instead of relying on open-source translators, we use DeepL, which provides high-quality translations. This decision is related to the low amount of available images (with respect to the original work of OpenAI). We wanted to avoid the risk of polluting the available data, however, we are aware of the bias (e.g., gender and age) that translation systems introduce during translations~\cite{hovy-etal-2020-sound}.

We translated 700K captions and we evaluated their quality. Some of the captions are available in Figure~\ref{tab:translations}.

To assess the quality of the translations, three of the authors (native Italian speakers) inspected a sample of 100 of the Italian translations, comparing them with the original English caption. They rated them with scores from 1 to 4. The meaning of the value is the following: 1, the sentence has lost its meaning, or it is not possible to understand it; 2, it is possible to get the idea but there is something wrong; 3, good, however, a native speaker might complain about some translations; 4, very good translation.

The average score was 3.78 suggesting that the translations were good on average. We also computed an inter-rater agreement with Gwet's AC1 using ordinal weighting obtaining a value of 0.858, suggesting a strong agreement between the annotators.

\begin{table*}[]
    \centering
    \begin{tabular}{l|l}\toprule
       \textbf{English Caption}  &  \textbf{Italian Caption} \\ \midrule
       \makecell[l]{an endless cargo of tanks on a train pulled  \\ down tracks in an empty dry landscape}   & \makecell[l]{un carico infinito di carri armati su un treno \\ trascinato lungo i binari in un paesaggio secco e vuoto} \\ \midrule
       person walking down the aisle & persona che cammina lungo la navata \\ \midrule
       popular rides at night at the county fair & giostre popolari di notte alla fiera della contea \\  \bottomrule
    \end{tabular}
    \caption{Translated captions from the Conceptual Captions dataset.}
    \label{tab:translations}
\end{table*}

\subsection{Data Cleaning}

Many of the captions in WIT describe ontological knowledge and encyclopedic facts (e.g., ``Roberto Baggio in 1994''). These kinds of texts are not useful to learn a good mapping between images and captions if additional information is not provided. To prevent polluting the data with captions that are not meaningful, we used Part-Of-Speech (POS) tagging on the text and removed all the captions that were composed for the 80\% or more by the tag PROPN, i.e. proper noun, (around ~10\% of the total captions for WIT). This simple solution allowed us to retain much of the dataset, without introducing noise. Captions like \textquotedblleft Dora Riparia'', \textquotedblleft Anna Maria Mozzoni'', \textquotedblleft Joey Ramone Place'', \textquotedblleft Kim Rhodes'', \textquotedblleft Ralph George Hawtrey'' which are proper nouns (PROPN) have been removed.

Regarding the dataset ILPOST, we used \texttt{langdetect}\footnote{\url{https://github.com/Mimino666/langdetect}} on its captions to remove non-Italian ones. Less than 2\% of captions got lost by this filter.

\section{Training}
\label{sec:training}

Our CLIP-Italian model is based on previous pre-trained state-of-the-art models for both the vision and textual part: we use Vision Transformer (ViT)~\cite{dosovitskiy2021an} and BERT~\cite{devlin-etal-2019-bert}. We limited the sequence length to a maximum of 96 tokens. We use a batch size of 128. For the optimization, we used AdaBelief~\cite{zhuang2020adabelief} with Adaptive Gradient Clipping (AGC) and a Cosine Annealing Schedule~\cite{loshchilov2016sgdr}. 

We applied augmentations to the images: we used random affine transformations and perspective changes, as well as occasional equalization and random changes to brightness, contrast, saturation, and hue. We made sure to keep hue augmentations limited to give the model the ability to learn color definitions.

Since we started from pre-trained checkpoints for both the vision and the text part, we found it useful to warm-up projection layers. To do so, we first train with both the vision and text encoder frozen until loss convergence. Then, we unfroze the rest of the model to fine-tune the entire architecture. We always pick the model that has the best evaluation loss, evaluating every 15 epochs.

The text encoder we use is the Italian BERT model\footnote{\url{https://huggingface.co/dbmdz/bert-base-italian-xxl-cased}} while for the vision encoder we use the pre-trained checkpoint provided by OpenAI.\footnote{\url{https://huggingface.co/openai/clip-vit-base-patch32}} Note that the OpenAI's vision transformer differs only for a few modifications from the original ViT.

Both images and texts are then projected to 512-dimensional vectors to which we apply the loss defined in CLIP using logit scaling equal to 20. We empirically observed that the logit scaling has a strong positive impact on performance of the model.

\section{Quantitative Evaluation}

To the best of our knowledge, CLIP-Italian is the first model of its kind. Hence, to provide meaningful comparisons, we use the multilingual CLIP model. This model was obtained through the use of multilingual knowledge distillation~\cite{reimers-gurevych-2020-making}. 

\subsection{Image Retrieval}

In the Image Retrieval task, we evaluate how well the images and their respective captions are mapped closely in the vector space. Given a caption, the task is to find the correct image that matches the caption from a set of images. This can be done by mapping a caption into the joint vector space and selecting the most similar image to it. We use the MSCOCO-IT validation dataset (that we have not used in the training set). The dataset contains 2000 images associated with their respective captions.

Table~\ref{tab:image_retrieval} presents the results for the image retrieval task.

\begin{table}[!t]
    \centering
    \begin{tabular}{l|cc} \toprule
       \textbf{Measure}  & \textbf{CLIP-Italian}  &  \textbf{mCLIP} \\ \bottomrule
         MRR@1 & \textbf{0.3797} & 0.2874 \\ 
         MRR@5 & \textbf{0.5039} & 0.3957 \\
         MRR@10 & \textbf{0.5204} &  0.4129 \\ \bottomrule 
    \end{tabular}
    \caption{Results on MSCOCO image retrieval task}
    \label{tab:image_retrieval}
\end{table}

\subsection{Zero-shot Classification}
The zero-shot image classification task replicates the experiment run by~\citet{CLIP21} on ImageNet. To do this, we used DeepL to automatically translate the image labels in ImageNet. We prepend all the labels in the test set with the article and translate them (\textit{a cat} is translated into ``un gatto''). The generated labels are then prepended with the text ``una foto di'' (a photo of) as in ``una foto di un gatto'' (a photo of a cat) to have the final prompt. Note that this is a much lighter prompt engineering than the one in~\citet{CLIP21} where different prompt templates are computed and averaged.
Given an input image and the so-generated prompts, we generate the embeddings (both for the image and all the prompts) and compute the similarities. We expect the correct label of the image to be the closest prompt in the embedding space. We compute accuracy metrics on these results.

Table~\ref{tab:zero_shot} presents the results for the zero-shot classification task. 

\begin{table}[!t]
    \centering
    \begin{tabular}{l|cc} \toprule
      \textbf{Measure} & \textbf{CLIP-Italian}  &  \textbf{mCLIP} \\ \bottomrule
       Accuracy@1  & \textbf{22.11} & 20.15 \\ 
       Accuracy@5 & \textbf{43.69} & 36.75 \\
       Accuracy@10 & \textbf{52.55} & 42.91 \\ \bottomrule
    \end{tabular}
    \caption{Results on ImageNet-1000 classification task}
    \label{tab:zero_shot}
\end{table}

\subsection{Discussion}

Our results confirm that CLIP-Italian is very competitive and outperforms mCLIP on the two different tasks we have been testing. Note that our results for zero-shot ImageNet classification of the CLIP-Italian model (trained on 1.4 million image-text pairs) are lower than those shown in \newcite{CLIP21} (trained on 400 million image-text pairs). However, considering that our results are in line with those obtained by mCLIP, we think that the translated image labels most probably had an impact on the final scores. Even with limited data sources, training time, and computational resources (two weeks of Google Cloud TPU sponsorship), the results obtained largely improve the state of the art in multi-modal image-text tasks for the Italian language.  

\section{Qualitative Evaluation}

Thanks to the support by HuggingFace, we are also able to provide an online demo that can be used to experiment with our model.\footnote{\url{https://huggingface.co/spaces/clip-italian/clip-italian-demo}}

\begin{figure}[h]
    \centering
    \includegraphics[width=\columnwidth]{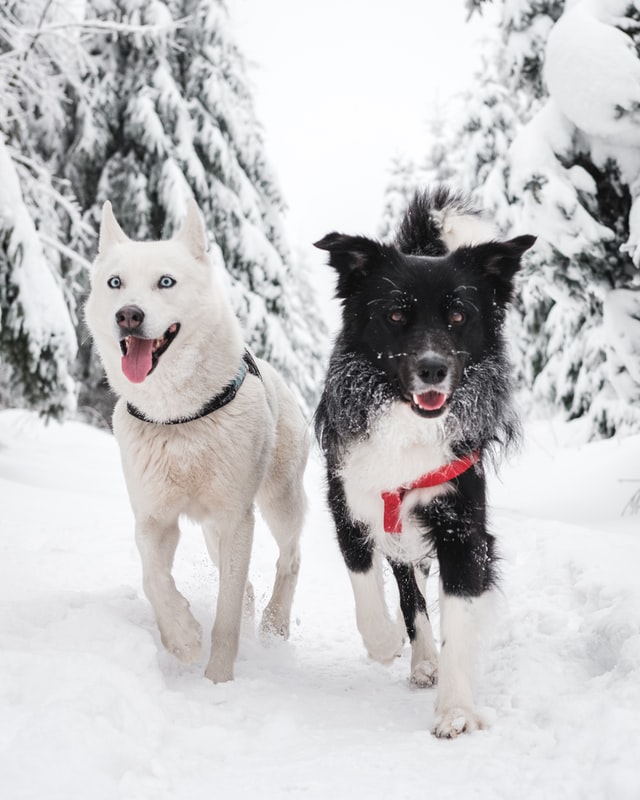}
    \caption{Result of the query ``due cani sulla neve'' (two dogs on the snow) on the Unsplash25K dataset.}
    \label{fig:qualitative:dogs}
\end{figure}

We show some examples related to the image retrieval task on the Unsplash25K dataset.\footnote{\url{https://github.com/unsplash/datasets}} Figure~\ref{fig:qualitative:dogs} shows the results of the query ``due cani sulla neve'' (two dogs on the snow), the model correctly finds the image, combining the concept of ``snow'' and the one of ``two dogs''. The model showed moderate numeracy capabilities during empirical evaluation, identifying up to three distinct or repeated elements inside images. We associate this behavior to implicit knowledge present in the training set, as we observe a steep drop in coherence when more than three elements are present. We believe this behavior is due to the likely low number of training points depicting more than three subjects in the scene. 

\begin{figure}[h]
    \centering
    \includegraphics[width=\columnwidth]{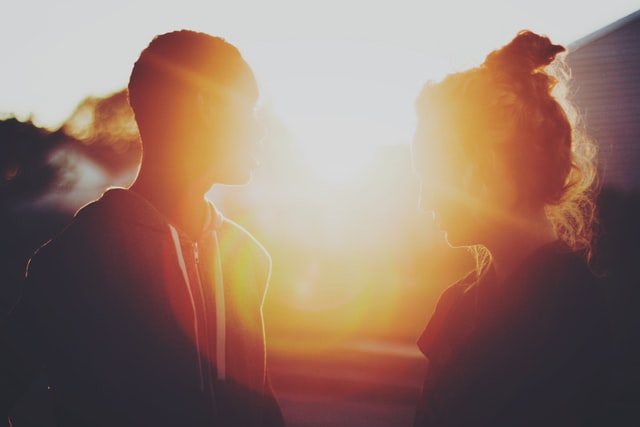}
    \caption{Result of the query ``una coppia al tramonto'' (a couple during sunset) on the Unsplash25K dataset.}
    \label{fig:qualitative:couple}
\end{figure}

\begin{figure}[h]
    \centering
    \includegraphics[width=\columnwidth]{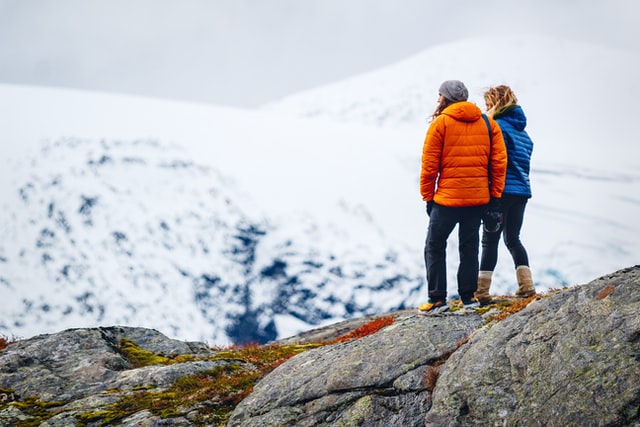}
    \caption{Result of the query ``una coppia in montagna'' (a couple on the mountain) on the Unsplash25K dataset.}
    \label{fig:qualitative:mountain}
\end{figure}

Figure~\ref{fig:qualitative:couple} shows similar performance for ``una coppia al tramonto'' (a couple during sunset), where the model was able to identify two people and the sun light in the background. A similar query, but with a mountain as a background, can be found in Figure~\ref{fig:qualitative:mountain}.

\begin{figure}[h]
    \centering
    \includegraphics[width=\columnwidth]{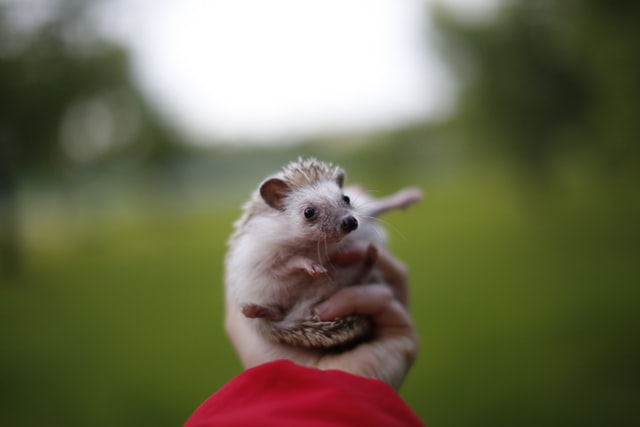}
    \caption{Result of the query ``un topolino'' (a small mouse) on the Unsplash25K dataset.}
    \label{fig:qualitative:topolino}
\end{figure}

However, the model is subject to limitations and biases. An example of a limitation is shown in Figure~\ref{fig:qualitative:topolino} where the query ``un topolino'' (a small mouse) resulted in finding an image of a small hedgehog. We also empirically found that the model shows some biases and has learned some stereotypes that need to be explored more in detail when using this model.

\section{Conclusions}

In this paper, we have presented the first CLIP model for the Italian language, trained on 1.4 million image-text pairs. The model shows promising performance in two well-established tasks, suggesting many possible future applications.

\section*{Ethical Considerations}
Large scale models are difficult and costly to train~\cite{bender2021dangers}; we computed the cost of the different experiments we ran and we estimated a total of 2,688\$ for each TPU used. This result comes from the hourly cost of the TPU (8\$) for 14 days; Note that we had access to a second TPU VM for part of the project and that in this estimate we are ignoring storage and data transfer costs. \newcite{strubell-etal-2019-energy} describe in detail how these model can have a huge environmental impact.

As \newcite{bianchi-hovy-2021-gap} describe, these computational needs are quickly becoming unfeasible for many universities, and even companies can struggle when applying large models on datasets, as also described by~\newcite{tagliabue-etal-2021-bert}.

\section*{Acknowledgments}
This work was only possible thanks to the Hugging Face and Google teams that provided two TPUs to train the CLIP-Italian model. Federico Bianchi is a member of the Bocconi Institute for Data Science and Analytics (BIDSA) and the Data and Marketing Insights (DMI) unit.

\bibliography{custom}
\bibliographystyle{acl_natbib}

\appendix

\section{Material Released}
\label{sec:appendix}

For the best transparency, and to allow further development upon our work, we release the pre-trained model,\footnote{\url{https://huggingface.co/clip-italian/clip-italian}} the modeling and training code,\footnote{\url{https://github.com/clip-italian/clip-italian}} a CometML report with training logs and metrics,\footnote{\url{https://www.comet.ml/g8a9/clip-italian/reports/clip-italian-training-metrics}}.
    
Interested readers can directly test CLIP-Italian's capabilities at our official demo.\footnote{\url{https://huggingface.co/spaces/clip-italian/clip-italian-demo}}

\end{document}